# Representation Learning on Graphs with Jumping Knowledge Networks

Keyulu Xu [1]  Chengtao Li [1]  Yonglong Tian [1]  Tomohiro Sonobe [2]
Ken-ichi Kawarabayashi [2]  Stefanie Jegelka [1]

## Abstract

Recent deep learning approaches for representation learning on graphs follow a neighborhood aggregation procedure. We analyze some important properties of these models, and propose a strategy to overcome those. In particular, the range of "neighboring" nodes that a node's representation draws from strongly depends on the graph structure, analogous to the spread of a random walk. To adapt to local neighborhood properties and tasks, we explore an architecture – jumping knowledge (JK) networks – that flexibly leverages, for each node, different neighborhood ranges to enable better structure-aware representation. In a number of experiments on social, bioinformatics and citation networks, we demonstrate that our model achieves state-of-the-art performance. Furthermore, combining the JK framework with models like Graph Convolutional Networks, GraphSAGE and Graph Attention Networks consistently improves those models' performance.

## 1. Introduction

Graphs are a ubiquitous structure that widely occurs in data analysis problems. Real-world graphs such as social networks, financial networks, biological networks and citation networks represent important rich information which is not seen from the individual entities alone, for example, the communities a person is in, the functional role of a molecule, and the sensitivity of the assets of an enterprise to external shocks. Therefore, representation learning of nodes in graphs aims to extract high-level features from a node as well as its neighborhood, and has proved extremely useful for many applications, such as node classification, clustering, and link prediction (Perozzi et al., 2014; Monti et al., 2017; Grover & Leskovec, 2016; Tang et al., 2015).

Recent works focus on deep learning approaches to node representation. Many of these approaches broadly follow a neighborhood aggregation (or "message passing" scheme), and those have been very promising (Kipf & Welling, 2017; Hamilton et al., 2017; Gilmer et al., 2017; Veličković et al., 2018; Kearnes et al., 2016). These models learn to iteratively aggregate the hidden features of every node in the graph with its adjacent nodes' as its new hidden features, where an iteration is parametrized by a layer of the neural network. Theoretically, an aggregation process of $k$ iterations makes use of the subtree structures of height $k$ rooted at every node. Such schemes have been shown to generalize the Weisfeiler-Lehman graph isomorphism test (Weisfeiler & Lehman, 1968) enabling to simultaneously learn the topology as well as the distribution of node features in the neighborhood (Shervashidze et al., 2011; Kipf & Welling, 2017; Hamilton et al., 2017).

Yet, such aggregation schemes sometimes lead to surprises. For example, it has been observed that the best performance with one of the state-of-the-art models, Graph Convolutional Networks (GCN), is achieved with a 2-layer model. Deeper versions of the model that, in principle, have access to more information, perform worse (Kipf & Welling, 2017). A similar degradation of learning for computer vision problems is resolved by residual connections (He et al., 2016a) that greatly aid the training of deep models. But, even with residual connections, GCNs with more layers do not perform as well as the 2-layer GCN on many datasets, e.g. citation networks.

Motivated by observations like the above, in this paper, we address two questions. First, we study properties and resulting limitations of neighborhood aggregation schemes. Second, based on this analysis, we propose an architecture that, as opposed to existing models, enables adaptive, *structure-aware* representations. Such representations are particularly interesting for representation learning on large complex graphs with diverse subgraph structures.

**Model analysis.** To better understand the behavior of different neighborhood aggregation schemes, we analyze the effective range of nodes that any given node's representation draws from. We summarize this sensitivity analysis by what we name the *influence distribution* of a node. This effective range implicitly encodes prior assumptions on what are the "nearest neighbors" that a node should draw information from. In particular, we will see that this influence is heavily affected by the graph structure, raising the question whether "one size fits all", in particular in graphs whose subgraphs have varying properties (such as more tree-like or more expander-like).

[1]Massachusetts Institute of Technology (MIT) [2]National Institute of Informatics, Tokyo. Correspondence to: Keyulu Xu <keyulu@mit.edu>, Stefanie Jegelka <stefje@mit.edu>.

In particular, our more formal analysis connects influence distributions with the spread of a random walk at a given node, a well-understood phenomenon as a function of the graph structure and eigenvalues (Lovász, 1993). For instance, in some cases and applications, a 2-step random walk influence that focuses on local neighborhoods can be more informative than higher-order features where some of the information may be "washed out" via averaging.

**Changing locality.** To illustrate the effect and importance of graph structure, recall that many real-world graphs possess locally strongly varying structure. In biological and citation networks, the majority of the nodes have few connections, whereas some nodes (hubs) are connected to many other nodes. Social and web networks usually consist of an expander-like core part and an almost-tree (bounded treewidth) part, which represent well-connected entities and the small communities respectively (Leskovec et al., 2009; Maehara et al., 2014; Tsonis et al., 2006).

Besides node features, this subgraph structure has great impact on the result of neighborhood aggregation. The speed of expansion or, equivalently, growth of the influence radius, is characterized by the random walk's mixing time, which changes dramatically on subgraphs with different structures (Lovász, 1993). Thus, the same number of iterations (layers) can lead to influence distributions of very different locality. As an example, consider the social network in Figure 1 from GooglePlus (Leskovec & Mcauley, 2012). The figure illustrates the expansions of a random walk starting at the square node. The walk (a) from a node within the core rapidly includes almost the entire graph. In contrast, the walk (b) starting at a node in the tree part includes only a very small fraction of all nodes. After 5 steps, the same walk has reached the core and, suddenly, spreads quickly. Translated to graph representation models, these spreads become the influence distributions or, in other words, the averaged features yield the new feature of the walk's starting node. This shows that in the same graph, the same number of steps can lead to very different effects. Depending on the application, wide-range or small-range feature combinations may be more desirable. A too rapid expansion may average too broadly and thereby lose information, while in other parts of the graph, a sufficient neighborhood may be needed for stabilizing predictions.

**JK networks.** The above observations raise the question whether it is possible to adaptively *adjust* (i.e., learn) the influence radii for each node and task. To achieve this, we explore an architecture that learns to selectively exploit information from neighborhoods of differing locality. This architecture selectively combines different aggregations at the last layer, i.e., the representations "jump" to the last layer. Hence, we name the resulting networks *Jumping Knowledge Networks (JK-Nets)*. We will see that empirically, when adaptation is an option, the networks indeed learn representations of different orders for different graph substructures. Moreover, in Section 6, we show that applying our framework to various state-of-the-art neighborhood-aggregation models consistently improves their performance.

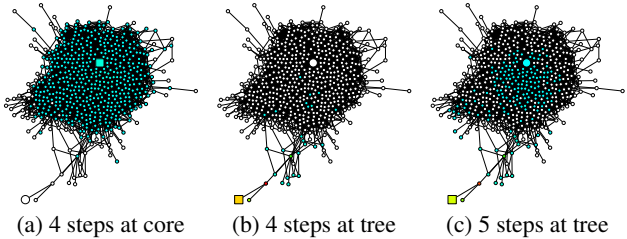

(a) 4 steps at core  (b) 4 steps at tree  (c) 5 steps at tree

*Figure 1.* Expansion of a random walk (and hence influence distribution) starting at (square) nodes in subgraphs with different structures. Different subgraph structures result in very different neighborhood sizes.

## 2. Background and Neighborhood aggregation schemes

We begin by summarizing some of the most common neighborhood aggregation schemes and, along the way, introduce our notation. Let $G = (V, E)$ be a simple graph with node features $X_v \in \mathbb{R}^{d_i}$ for $v \in V$. Let $\widetilde{G}$ be the graph obtained by adding a self-loop to every $v \in V$. The hidden feature of node $v$ learned by the $l$-th layer of the model is denoted by $h_v^{(l)} \in \mathbb{R}^{d_h}$. Here, $d_i$ is the dimension of the input features and $d_h$ is the dimension of the hidden features, which, for simplicity of exposition, we assume to be the same across layers. We also use $h_v^{(0)} = X_v$ for the node feature. The neighborhood $N(v) = \{u \in V \mid (v, u) \in E\}$ of node $v$ is the set of adjacent nodes of $v$. The analogous neighborhood $\widetilde{N}(v) = \{v\} \cup \{u \in V \mid (v, u) \in E\}$ on $\widetilde{G}$ includes $v$.

A typical neighborhood aggregation scheme can generically be written as follows: for a $k$-layer model, the $l$-th layer ($l = 1..k$) updates $h_v^{(l)}$ for every $v \in V$ simultaneously as

$$h_v^{(l)} = \sigma \left( W_l \cdot \text{AGGREGATE}\left( \{h_u^{(l-1)}, \forall u \in \widetilde{N}(v)\} \right) \right) \quad (1)$$

where AGGREGATE is an aggregation function defined by the specific model, $W_l$ is a trainable weight matrix on the $l$-th layer shared by all nodes, and $\sigma$ is a non-linear activation function, e.g. a ReLU.

**Graph Convolutional Networks (GCN).** Graph Convolutional Networks (GCN) (Kipf & Welling, 2017), initially motivated by spectral graph convolutions (Hammond et al., 2011; Defferrard et al., 2016), are a specific instantiation of this framework (Gilmer et al., 2017), of the form

$$h_v^{(l)} = \text{ReLU}\Big(W_l \cdot \sum_{u \in \widetilde{N}(v)} (\deg(v)\deg(u))^{-1/2} h_u^{(l-1)}\Big) \tag{2}$$

where $\deg(v)$ is the degree of node $v$ in $G$. Hamilton et al. (2017) derived a variant of GCN that also works in inductive settings (previously unseen nodes), by using a different normalization to average:

$$h_v^{(l)} = \text{ReLU}\Big(W_l \cdot \frac{1}{\widetilde{\deg}(v)} \sum_{u \in \widetilde{N}(v)} h_u^{(l-1)}\Big) \tag{3}$$

where $\widetilde{\deg}(v)$ is the degree of node $v$ in $\widetilde{G}$.

**Neighborhood Aggregation with Skip Connections.** Instead of aggregating a node and its neighbors at the same time as in Eqn. (1), a number of recent approaches aggregate the neighbors first and then combine the resulting neighborhood representation with the node's representation from the last iteration. More formally, each node is updated as

$$h_{N(v)}^{(l)} = \sigma\left(W_l \cdot \text{AGGREGATE}_N\big(\{h_u^{(l-1)}, \forall u \in N(v)\}\big)\right)$$
$$h_v^{(l)} = \text{COMBINE}\left(h_v^{(l-1)}, h_{N(v)}^{(l)}\right)$$

where $\text{AGGREGATE}_N$ and COMBINE are defined by the specific model. The COMBINE step is key to this paradigm and can be viewed as a form of a "skip connection" between different layers. For COMBINE, GraphSAGE (Hamilton et al., 2017) uses concatenation after a feature transform. Column Networks (Pham et al., 2017) interpolate the neighborhood representation and the node's previous representation, and Gated GNN (Li et al., 2016) uses the Gated Recurrent Unit (GRU) (Cho et al., 2014). Another well-known variant of skip connections, residual connections, use the identity mapping to help signals propagate (He et al., 2016a;b).

These skip connections are input- but not output-unit specific: If we "skip" a layer for $h_v^{(l)}$ (do not aggregate) or use a certain COMBINE, all subsequent units using this representation will be using this skip implicitly. It is impossible that a certain higher-up representation $h_u^{(l+j)}$ uses the skip and another one does not. As a result, skip connections cannot adaptively adjust the neighborhood sizes of the final-layer representations independently.

**Neighborhood Aggregation with Directional Biases.** Some recent models, rather than treating the features of adjacent nodes equally, weigh "important" neighbors more. This paradigm can be viewed as neighborhood-aggregation with directional biases because a node will be influenced by some directions of expansion more than the others.

Graph Attention Networks (GAT) (Veličković et al., 2018) and VAIN (Hoshen, 2017) learn to select the important neighbors via an attention mechanism. The max-pooling operation in GraphSAGE (Hamilton et al., 2017) implicitly selects the important nodes. This line of work is orthogonal to ours, because it modifies the direction of expansion whereas our model operates on the locality of expansion. Our model can be combined with these models to add representational power. In Section 6, we demonstrate that our framework works with not only simple neighborhood-aggregation models (GCN), but also with skip connections (GraphSAGE) and directional biases (GAT).

## 3. Influence Distribution and Random Walks

Next, we explore some important properties of the above aggregation schemes. Related to ideas of sensitivity analysis and influence functions in statistics (Koh & Liang, 2017) that measure the influence of a training point on parameters, we study the range of nodes whose features affect a given node's representation. This range gives insight into how large a neighborhood a node is drawing information from.

We measure the sensitivity of node $x$ to node $y$, or the influence of $y$ on $x$, by measuring how much a change in the input feature of $y$ affects the representation of $x$ in the last layer. For any node $x$, the *influence distribution* captures the relative influences of all other nodes.

**Definition 3.1** (Influence score and distribution). *For a simple graph $G = (V, E)$, let $h_x^{(0)}$ be the input feature and $h_x^{(k)}$ be the learned hidden feature of node $x \in V$ at the $k$-th (last) layer of the model. The influence score $I(x, y)$ of node $x$ by any node $y \in V$ is the sum of the absolute values of the entries of the Jacobian matrix $\left[\frac{\partial h_x^{(k)}}{\partial h_y^{(0)}}\right]$. We define the influence distribution $I_x$ of $x \in V$ by normalizing the influence scores: $I_x(y) = I(x, y) / \sum_z I(x, z)$, or*

$$I_x(y) = e^T \left[\frac{\partial h_x^{(k)}}{\partial h_y^{(0)}}\right] e \Bigg/ \left(\sum_{z \in V} e^T \left[\frac{\partial h_x^{(k)}}{\partial h_z^{(0)}}\right] e\right)$$

*where $e$ is the all-ones vector.*

Later, we will see connections of influence distributions with random walks. For completeness, we also define random walk distributions.

**Definition 3.2.** *Consider a random walk on $\widetilde{G}$ starting at a node $v_0$; if at the $t$-th step we are at a node $v_t$, we move to any neighbor of $v_t$ (including $v_t$) with equal probability.*

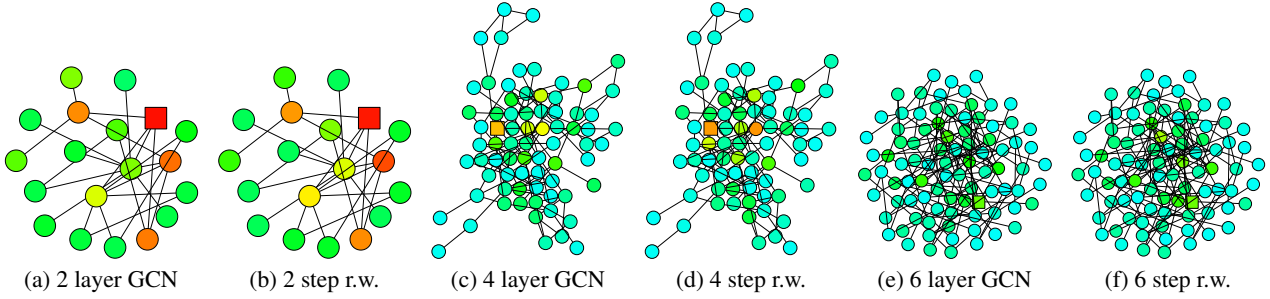

Figure 2. Influence distributions of GCNs and random walk distributions starting at the square node

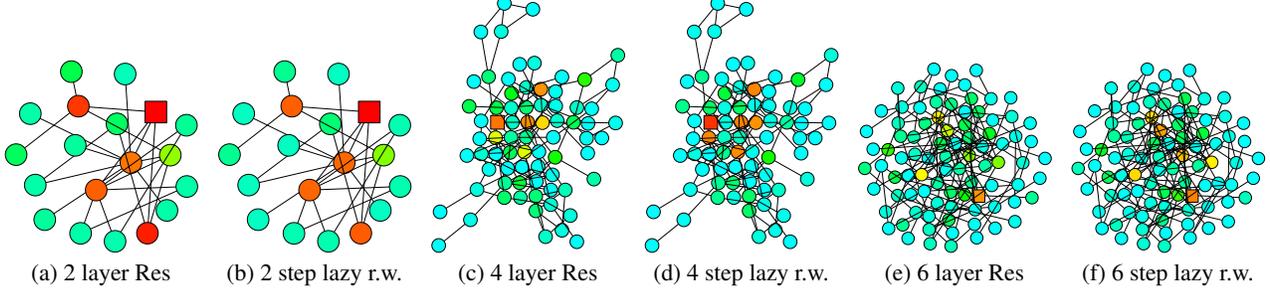

Figure 3. Influence distributions of GCNs with residual connections and random walk distributions with lazy factor 0.4

*The $t$-step random walk distribution $P_t$ of $v_0$ is*

$$P_t(i) = Prob(v_t = i). \qquad (4)$$

*Analogous definitions apply for random walks with non-uniform transition probabilities.*

An important property of the random walk distribution is that it becomes more spread out as $t$ increases and converges to the limit distribution if the graph is non-bipartite. The rate of convergence depends on the structure of the subgraph and can be bounded by the spectral gap (or the conductance) of the random walk's transition matrix (Lovász, 1993).

### 3.1. Model Analysis

The influence distribution for different aggregation models and nodes can give insights into the information captured by the respective representations. The following results show that the influence distributions of common aggregation schemes are closely connected to random walk distributions. This observation hints at specific implications – strengths and weaknesses – that we will discuss.

With a randomization assumption of the ReLU activations similar to that in (Kawaguchi, 2016; Choromanska et al., 2015), we can draw connections between GCNs and random walks:

**Theorem 1.** *Given a $k$-layer GCN with averaging as in Equation* (3)*, assume that all paths in the computation graph of the model are activated with the same probability of success $\rho$. Then the influence distribution $I_x$ for any node $x \in V$ is equivalent, in expectation, to the $k$-step random walk distribution on $\widetilde{G}$ starting at node $x$.*

We prove Theorem 1 in the appendix.

It is straightforward to modify the proof of Theorem 1 to show a nearly equivalent result for the version of GCN in Equation (2). The only difference is that each random walk path $v_p^0, v_p^1, ..., v_p^k$ from node $x$ ($v_p^0$) to $y$ ($v_p^k$), instead of probability $\rho \prod_{l=1}^{k} \frac{1}{\widetilde{\deg}(v_p^l)}$, now has probability $\frac{\rho}{Q} \prod_{l=1}^{k-1} \frac{1}{\widetilde{\deg}(v_p^l)} \cdot (\widetilde{\deg}(x)\widetilde{\deg}(y))^{-1/2}$, where $Q$ is a normalizing factor. Thus, the difference in probability is small, especially when the degree of $x$ and $y$ are close.

Similarly, we can show that neighborhood aggregation schemes with directional biases resemble biased random walk distributions. This follows by substituting the corresponding probabilities into the proof of Theorem 1.

Empirically, we observe that, despite somewhat simplifying assumptions, our theory is close to what happens in practice. We visualize the heat maps of the influence distributions for a node (labeled square) for trained GCNs, and compare with the random walk distributions starting at the same node. Figure 2 shows example results. Darker colors correspond to higher influence probabilities. To show the effect of skip connections, Figure 3 visualizes the analogous heat maps for one example—GCN with residual connections. Indeed, we observe that the influence distributions of networks with residual connections approximately correspond to lazy random walks: each step has a higher probability of staying at the current node. Local information is retained with similar probabilities for all nodes in each iteration; this cannot adapt to diverse needs of specific upper-layer nodes. Further visualizations may be found in the appendix.

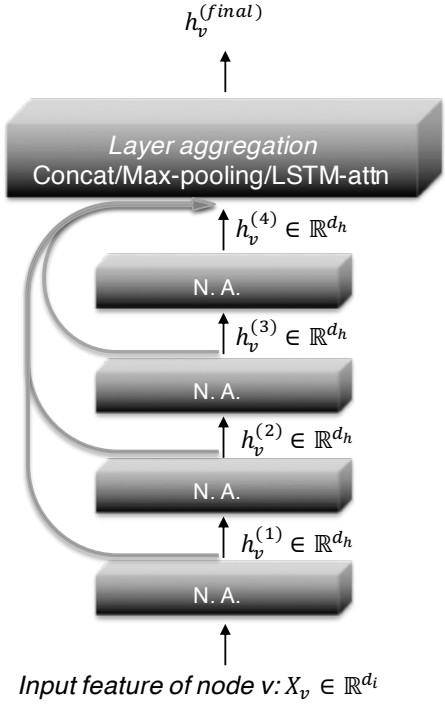

*Figure 4.* A 4-layer Jumping Knowledge Network (JK-Net). N.A. stands for neighborhood aggregation.

**Fast Collapse on Expanders.** To better understand the implication of Theorem 1 and the limitations of the corresponding neighborhood aggregation algorithms, we revisit the scenario of learning on a social network shown in Figure 1. Random walks starting inside an expander converge rapidly in $O(\log |V|)$ steps to an almost-uniform distribution (Hoory et al., 2006). After $O(\log |V|)$ iterations of neighborhood aggregation, by Theorem 1 the representation of every node is influenced almost equally by any other node in the expander. Thus, the node representations will be representative of the global graph and carry limited information about individual nodes. In contrast, random walks starting at the bounded tree-width (almost-tree) part converge slowly, i.e., the features retain more local information. Models that impose a fixed random walk distribution inherit these discrepancies in the speed of expansion and influence neighborhoods, which may not lead to the best representations for all nodes.

## 4. Jumping Knowledge Networks

The above observations raise the question whether the fixed but structure-dependent influence radius size induced by common aggregation schemes really achieves the best representations for all nodes and tasks. Large radii may lead to too much averaging, while small radii may lead to instabilities or insufficient information aggregation. Hence, we propose two simple yet powerful architectural changes – jump connections and a subsequent selective but adaptive aggregation mechanism.

Figure 4 illustrates the main idea: as in common neighborhood aggregation networks, each layer increases the size of the influence distribution by aggregating neighborhoods from the previous layer. At the last layer, for each node, we carefully select from all of those itermediate representations (which "jump" to the last layer), potentially combining a few. If this is done independently for each node, then the model can adapt the effective neighborhood size for each node as needed, resulting in exactly the desired adaptivity.

Our model permits general layer-aggregation mechanisms. We explore three approaches; others are possible too. Let $h_v^{(1)}, ..., h_v^{(k)}$ be the jumping representations of node $v$ (from $k$ layers) that are to be aggregated.

**Concatenation.** A concatenation $\left[h_v^{(1)}, ..., h_v^{(k)}\right]$ is the most straightforward way to combine the layers, after which we may perform a linear transformation. If the transformation weights are shared across graph nodes, this approach is not node-adaptive. Instead, it optimizes the weights to combine the subgraph features in a way that works best for the dataset overall. One may expect concatenation to be suitable for small graphs and graphs with regular structure that require less adaptivity; also because weight-sharing helps reduce overfitting.

**Max-pooling.** An element-wise $\max\left(h_v^{(1)}, ..., h_v^{(k)}\right)$ selects the most informative layer *for each feature coordinate*. For example, feature coordinates that represent more local properties can use the feature coordinates learned from the close neighbors and those representing global status would favor features from the higher-up layers. Max-pooling is adaptive and has the advantage that it does not introduce any additional parameters to learn.

**LSTM-attention.** An attention mechanism identifies the most useful neighborhood ranges for each node $v$ by computing an attention score $s_v^{(l)}$ for each layer $l$ $\left(\sum_l s_v^{(l)} = 1\right)$, which represents the importance of the feature learned on the $l$-th layer for node $v$. The aggregated representation for node $v$ is a weighted average of the layer features $\sum_l s_v^{(l)} \cdot h_v^{(l)}$. For LSTM attention, we input $h_v^{(1)}, ..., h_v^{(k)}$ into a bi-directional LSTM (Hochreiter & Schmidhuber, 1997) and generate the forward-LSTM and backward-LSTM hidden features $f_v^{(l)}$ and $b_v^{(l)}$ for each layer $l$. A linear mapping of the concatenated features $[f_v^{(l)} || b_v^{(l)}]$ yields the scalar importance score $s_v^{(l)}$. A Softmax layer applied to $\{s_v^{(l)}\}_{l=1}^k$ yields the attention of node $v$ on its neighborhood in different ranges. Finally we take the sum of $[f_v^{(l)}||b_v^{(l)}]$ weighted by $\texttt{SoftMax}(\{s_v^{(l)}\}_{l=1}^k)$ to get the final layer representation. Another possible implementation combines LSTM with max-pooling. LSTM-attention is node adaptive because the attention scores are different for each node. We shall see that the this approach shines on large complex graphs, although it may overfit on small graphs (fewer training nodes) due to its relatively higher complexity.

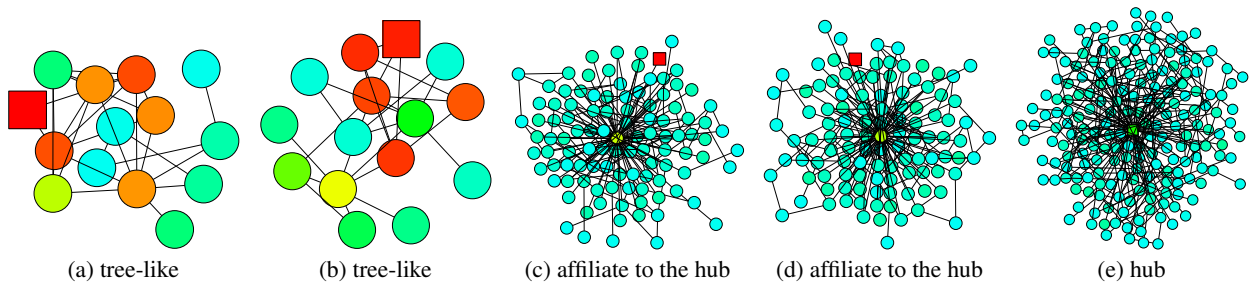

(a) tree-like  (b) tree-like  (c) affiliate to the hub  (d) affiliate to the hub  (e) hub

*Figure 5.* A 6-layer JK-Net learns to adapt to different subgraph structures

### 4.1. JK-Net Learns to Adapt

The key idea for the design of layer-aggregation functions is to determine the importance of a node's subgraph features at different ranges after looking at the learned features on all layers, rather than to optimize and fix the same weights for all nodes. Under the same assumption on the ReLU activation distribution as in Theorem 1, we show below that layer-wise max-pooling implicitly learns the influence locality adaptively for different nodes. The proof for layer-wise attention follows similarly.

**Proposition 1.** *Assume that paths of the same length in the computation graph are activated with the same probability. The influence score $I(x, y)$ for any $x, y \in V$ under a $k$-layer JK-Net with layer-wise max-pooling is equivalent in expectation to a mixture of $0, .., k$-step random walk distributions on $\widetilde{G}$ at $y$ starting at $x$, the coefficients of which depend on the values of the layer features $h_x^{(l)}$.*

We prove Proposition 1 in the appendix. Contrasting this result with the influence distributions of other aggregation mechanisms, we see that JK-networks indeed differ in their node-wise adaptivity of neighborhood ranges.

Figure 5 illustrates how a 6-layer JK-Net with max-pooling aggregation learns to adapt to different subgraph structures on a citation network. Within a tree-like structure, the influence stays in the "small community" the node belongs to. In contrast, 6-layer models whose influence distributions follow random walks, e.g. GCNs, would reach out too far into irrelevant parts of the graph, and models with few layers may not be able to cover the entire "community", as illustrated in Figure 1, and Figures 7, 8 in the appendix. For a node affiliated to a "hub", which presumably plays the role of connecting different types of nodes, JK-Net learns to put most influence on the node itself and otherwise spreads out the influence. GCNs, however, would not capture the importance of the node's own features in such a structure because the probability at an affiliate node is small after a few random walk steps. For hubs, JK-Net spreads out the influence across the neighboring nodes in a reasonable range, which makes sense because the nodes connected to the hubs are presumably as informative as the hubs' own features. For comparison, Table 6 in the appendix includes more visualizations of how models with random walk priors behave.

### 4.2. Intermediate Layer Aggregation and Structures

Looking at Figure 4, one may wonder whether the same inter-layer connections could be drawn between all layers. The resulting architecture is approximately a graph correspondent of DenseNets, which were introduced for computer vision problems (Huang et al., 2017), if the layer-wise concatenation aggregation is applied. This version, however, would require many more features to learn. Viewing the DenseNet setting (images) from a graph-theoretic perspective, images correspond to regular, in fact, near-planar graphs. Such graphs are far from being expanders, and do not pose the challenges of graphs with varying subgraph structures. Indeed, as we shall see, models with concatenation aggregation perform well on graphs with more regular structures such as images and well-structured communities. As a more general framework, JK-Net admits general layer-wise aggregation models and enables better structure-aware representations on graphs with complex structures.

## 5. Other Related Work

Spectral graph convolutional neural networks apply convolution on graphs by using the graph Laplacian eigenvectors as the Fourier atoms (Bruna et al., 2014; Shuman et al., 2013; Defferrard et al., 2016). A major drawback of the spectral methods, compared to spatial approaches like neighborhood-aggregation, is that the graph Laplacian needs to be known in advance. Hence, they cannot generalize to unseen graphs.

| Dataset  | Nodes   | Edges       | Classes | Features |
|----------|---------|-------------|---------|----------|
| Citeseer | 3,327   | 4,732       | 6       | 3,703    |
| Cora     | 2,708   | 5,429       | 7       | 1,433    |
| Reddit   | 232,965 | avg deg 492 | 50      | 300      |
| PPI      | 56,944  | 818,716     | 121     | 50       |

Table 1. Dataset statistics

| Model         | Citeseer   | Model         | Cora       |
|---------------|------------|---------------|------------|
| GCN (2)       | 77.3 (1.3) | GCN (2)       | 88.2 (0.7) |
| GAT (2)       | 76.2 (0.8) | GAT (3)       | 87.7 (0.3) |
| JK-MaxPool (1)| 77.7 (0.5) | JK-Maxpool (6)| **89.6** (0.5) |
| JK-Concat (1) | **78.3** (0.8) | JK-Concat (6) | 89.1 (1.1) |
| JK-LSTM (2)   | 74.7 (0.9) | JK-LSTM (1)   | 85.8 (1.0) |

Table 2. Results of GCN-based JK-Nets on Citeseer and Cora. The baselines are GCN and GAT. The number in parentheses next to the model name indicates the best-performing number of layers among 1 to 6. Accuracy and standard deviation are computed from 3 random data splits.

## 6. Experiments

We evaluate JK-Nets on four benchmark datasets. (I) The task on citation networks (Citeseer, Cora) (Sen et al., 2008) is to classify academic papers into different subjects. The dataset contains bag-of-words features for each document (node) and citation links (edges) between documents. (II) On Reddit (Hamilton et al., 2017), the task is to predict the community to which different Reddit posts belong. Reddit is an online discussion forum where users comment in different topical communities. Two posts (nodes) are connected if some user commented on both posts. The dataset contains word vectors as node features. (III) For protein-protein interaction networks (PPI) (Hamilton et al., 2017), the task is to classify protein functions. PPI consists of 24 graphs, each corresponds to a human tissue. Each node has positional gene sets, motif gene sets and immunological signatures as features and gene ontology sets as labels. 20 graphs are used for training, 2 graphs are used for validation and the rest for testing. Statistics of the datasets are summarized in Table 1.

**Settings.** In the *transductive setting*, we are only allowed to access a subset of nodes in one graph as training data, and validate/test on others. Our experiments on Citeseer, Cora and Reddit are transductive. In the *inductive setting*, we use a number of full graphs as training data and use other completely unseen graphs as validation/testing data. Our experiments on PPI are inductive.

We compare against three baselines: Graph Convolutional Networks (GCN) (Kipf & Welling, 2017), GraphSAGE (Hamilton et al., 2017) and Graph Attention Networks (GAT) (Veličković et al., 2018).

### 6.1. Citeseer & Cora

For experiments on Citeseer and Cora, we choose GCN as the base model since on our data split, it is outperforming GAT. We construct JK-Nets by choosing MaxPooling (JK-MaxPool), Concatenation (JK-Concat), or LSTM-attention (JK-LSTM) as final aggregation layer. When taking the final aggregation, besides normal graph convolutional layers, we also take the first linear-transformed representation into account. The final prediction is done via a fully connected layer on top of the final aggregated representation. We split nodes in each graph into 60%, 20% and 20% for training, validation and testing. We vary the number of layers from 1 to 6 for each model and choose the best performing model with respect to the validation set. Throughout the experiments, we use the Adam optimizer (Kingma & Ba, 2014) with learning rate 0.005. We fix the dropout rate to be 0.5, the dimension of hidden features to be within $\{16, 32\}$, and add an $L2$ regularization of 0.0005 on model parameters. The results are shown in Table 2.

**Results.** We observe in Table 2 that JK-Nets outperform both GCN and GAT baselines in terms of prediction accuracy. Though JK-Nets perform well in general, there is no consistent winner and performance varies slightly across datasets.

Taking a closer look at results on Cora, both GCN and GAT achieve their best accuracies with only 2 or 3 layers, suggesting that local information is a stronger signal for classification than global ones. However, the fact that JK-Nets achieve the best performance with 6 layers indicates that global together with local information will help boost performance. This is where models like JK-Nets can be particularly beneficial. LSTM-attention may not be suitable for such small graphs because of its relatively high complexity.

### 6.2. Reddit

The Reddit data is too large to be handled well by current implementations of GCN or GAT. Hence, we use the more scalable GraphSAGE as the base model for JK-Net. It has skip connections and different modes of node aggregation. We experiment with Mean and MaxPool node aggregators, which take mean and max-pooling of a *linear transformation* of representations of the sampled neighbors. Combining each of GraphSAGE modes with MaxPooling, Concatenation or LSTM-attention as the last aggregation layer gives 6 JK-Net variants. We follow exactly the same setting of GraphSAGE as in the original paper (Hamilton et al., 2017), where the model consists of 2 hidden layers, each with 128 hidden units and is trained with Adam with learning rate of 0.01 and no weight decay. Results are shown in Table 3.

**Results.** With MaxPool as node aggregator and Concat as layer aggregator, JK-Net achieves the best Micro-F1 score

Representation Learning on Graphs with Jumping Knowledge Networks| JK \ Node | GraphSAGE | Maxpool | Concat | LSTM |
|---|---|---|---|---|
| Mean | 0.950 | 0.953 | 0.955 | 0.950 |
| MaxPool | 0.948 | 0.924 | **0.965** | 0.877 |

Table 3. Results of GraphSAGE-based JK-Nets on Reddit. The baseline is GraphSAGE. Model performance is measured in Micro-F1 score. Each column shows the results of a JK-Net variant. For all models, the number of layers is fixed to 2.

| JK \ Node | SAGE | MaxPool | Concat | LSTM |
|---|---|---|---|---|
| Mean (10 epochs) | 0.644 | 0.658 | 0.667 | **0.721** |
| Mean (30 epochs) | 0.690 | 0.713 | 0.694 | **0.818** |
| MaxPool (10 epochs) | 0.668 | 0.671 | 0.687 | 0.621* |

Table 4. Results of GraphSAGE-based JK-Net on the PPI data. The baseline is GraphSAGE (SAGE). Each column, excluding SAGE, represents a JK-Net with different layer aggregation. All models use 3 layers, except for those with "*", whose number of layers is set to 2 due to GPU memory constraints. 0.6 is the corresponding 2-layer GraphSAGE performance.

among GarphSAGE and JK-Net variants. Note that the original GraphSAGE already performs fairly well with a Micro-F1 of 0.95. JK-Net reduces the error by 30%. The communities in the Reddit dataset were explicitly chosen from the well-behaved middle-sized communities to avoid the noisy cores and tree-like small communities (Hamilton et al., 2017). As a result, this graph is more regular than the original Reddit data, and hence not exhibit the problems of varying subgraph structures. In such a case, the added flexibility of the node-specific neighborhood choices may not be as relevant, and the stabilizing properties of concatenation instead come into play.

### 6.3. PPI

We demonstrate the power of adaptive JK-Nets, e.g., JK-LSTM, with experiments on the PPI data, where the subgraphs have more diverse and complex structures than those in the Reddit community detection dataset. We use both GraphSAGE and GAT as base models for JK-Net. The implementation of GraphSAGE and GAT are quite different: GraphSAGE is sample-based, where neighbors of a node are sampled to be a fixed number, while GAT considers all neighbors. Such differences cause large gaps in terms of both scalability and performances. Given that GraphSAGE scales to much larger graphs, it appears particularly valuable to evaluate how much JK-Net can improve upon GraphSAGE.

For GraphSAGE we follow the setup as in the Reddit experiments, except that we use 3 layers when possible, and compare the performance after 10 and 30 epochs of training. The results are shown in Table 4. For GAT and its JK-Net variants we stack two hidden layers with 4 attention heads computing 256 features (for a total of 1024 features), and a final prediction layer with 6 attention heads computing 121 features each. They are further averaged and input into sigmoid activations. We employ skip connections across intermediate attentional layers. These models are trained with Batch-size 2 and Adam optimizer with learning rate of 0.005. The results are shown in Table 5.

**Results.** JK-Nets with the LSTM-attention aggregators outperform the non-adaptive models GraphSAGE, GAT and JK-Nets with concatenation aggregators. In particular, JK-LSTM outperforms GraphSAGE by 0.128 in terms of micro-

| Model | PPI |
|---|---|
| MLP | 0.422 |
| GAT | 0.968 (0.002) |
| JK-Concat (2) | 0.959 (0.003) |
| JK-LSTM (3) | 0.969 (0.006) |
| JK-Dense-Concat (2)* | 0.956 (0.004) |
| JK-Dense-LSTM (2)* | **0.976** (0.007) |

Table 5. Micro-F1 scores of GAT-based JK-Nets on the PPI data. The baselines are GAT and MLP (Multilayer Perceptron). While the number of layers for JK-Concat and JK-LSTM are chosen from {2, 3}, the ones for JK-Dense-Concat and JK-Dense-LSTM are directly set to 2 due to GPU memory constraints.

F1 score after 30 epochs of training. Structure-aware node adaptive models are especially beneficial on such complex graphs with diverse structures.

## 7. Conclusion

Motivated by observations that reveal great differences in neighborhood information ranges for graph node embeddings, we propose a new aggregation scheme for node representation learning that can adapt neigborhood ranges to nodes individually. This JK-network can improve representations in particular for graphs that have subgraphs of diverse local structure, and may hence not be well captured by fixed numbers of neighborhood aggregations. Interesting directions for future work include exploring other layer aggregators and studying the effect of the combination of various layer-wise and node-wise aggregators on different types of graph structures.

## Acknowledgements

This research was supported by NSF CAREER award 1553284, and JST ERATO Kawarabayashi Large Graph Project, Grant Number JPMJER1201, Japan.## References

Bruna, J., Zaremba, W., Szlam, A., and LeCun, Y. Spectral networks and locally connected networks on graphs.

Representation Learning on Graphs with Jumping Knowledge NetworksPham, T., Tran, T., Phung, D. Q., and Venkatesh, S. Column networks for collective classification. In *AAAI Conference on Artificial Intelligence*, pp. 2485–2491, 2017.

Sen, P., Namata, G., Bilgic, M., Getoor, L., Galligher, B., and Eliassi-Rad, T. Collective classification in network data. *AI magazine*, 29(3):93, 2008.

Shervashidze, N., Schweitzer, P., Leeuwen, E. J. v., Mehlhorn, K., and Borgwardt, K. M. Weisfeiler-lehman graph kernels. *Journal of Machine Learning Research*, 12(Sep):2539–2561, 2011.

Shuman, D. I., Narang, S. K., Frossard, P., Ortega, A., and Vandergheynst, P. The emerging field of signal processing on graphs: Extending high-dimensional data analysis to networks and other irregular domains. *IEEE Signal Processing Magazine*, 30(3):83–98, 2013.

Tang, J., Qu, M., Wang, M., Zhang, M., Yan, J., and Mei, Q. Line: Large-scale information network embedding. In *Proceedings of the International World Wide Web Conference (WWW)*, pp. 1067–1077, 2015.

Tsonis, A. A., Swanson, K. L., and Roebber, P. J. What do networks have to do with climate? *Bulletin of the American Meteorological Society*, 87(5):585–595, 2006.

Veličković, P., Cucurull, G., Casanova, A., Romero, A., Liò, P., and Bengio, Y. Graph attention networks. *International Conference on Learning Representations (ICLR)*, 2018.

Weisfeiler, B. and Lehman, A. A reduction of a graph to a canonical form and an algebra arising during this reduction. *Nauchno-Technicheskaya Informatsia*, 2(9): 12–16, 1968.

## A. Proof for Theorem 1

*Proof.* Denote by $f_x^{(l)}$ the pre-activated feature of $h_x^{(l)}$, i.e. $\frac{1}{\widetilde{\deg}(x)} \cdot \sum_{z \in \widetilde{N}(x)} W_l h_z^{(l-1)}$, for any $l = 1..k$, we have

$$\frac{\partial h_x^{(l)}}{\partial h_y^{(0)}} = \frac{1}{\widetilde{\deg}(x)} \cdot \text{diag}\left(1_{f_x^{(l)} > 0}\right) \cdot W_l \cdot \sum_{z \in \widetilde{N}(x)} \frac{\partial h_z^{(l-1)}}{\partial h_y^{(0)}}$$

By chain rule, we get

$$\frac{\partial h_x^{(k)}}{\partial h_y^{(0)}} = \sum_{p=1}^{\Psi} \left[\frac{\partial h_x^{(k)}}{\partial h_y^{(0)}}\right]_p$$

$$= \sum_{p=1}^{\Psi} \prod_{l=k}^{1} \frac{1}{\widetilde{\deg}(v_p^l)} \cdot \text{diag}\left(1_{f_{v_p^l}^{(l)} > 0}\right) \cdot W_l$$

Here, $\Psi$ is the total number of paths $v_p^k v_p^{k-1}, ..., v_p^1, v_p^0$ of length $k+1$ from node $x$ to node $y$. For any path $p$, $v_p^k$ is node $x$, $v_p^0$ is node $y$ and for $l = 1..k-1$, $v_p^{l-1} \in \widetilde{N}(v_p^l)$.

As for each path $p$, the derivative $\left[\frac{\partial h_x^{(k)}}{\partial h_y^{(0)}}\right]_p$ represents a directed acyclic computation graph, where the input neurons are the same as the entries of $W_1$, and at a layer $l$. We can express an entry of the derivative as

$$\left[\frac{\partial h_x^{(k)}}{\partial h_y^{(0)}}\right]_p^{(i,j)} = \prod_{l=k}^{1} \frac{1}{\widetilde{\deg}(v_p^l)} \sum_{q=1}^{\Phi} Z_q \prod_{l=k}^{1} w_q^{(l)}$$

Here, $\Phi$ is the number of paths $q$ from the input neurons to the output neuron $(i, j)$, in the computation graph of $\left[\frac{\partial h_x^{(k)}}{\partial h_y^{(0)}}\right]_p$. For each layer $l$, $w_q^l$ is the entry of $W_l$ that is used in the $q$-th path. Finally, $Z_q \in \{0, 1\}$ represents whether the $q$-th path is active ($Z_q = 1$) or not ($Z_q = 0$) as a result of the ReLU activation of the entries of $f_{v_p^l}^{(l)}$'s on the $q$-th path.

Under the assumption that the $Z$'s are Bernoulli random variables with the same probability of success, for all $q$, $\Pr(Z_q = 1) = \rho$, we have $\mathbb{E}\left[\left[\frac{\partial h_x^{(k)}}{\partial h_y^{(0)}}\right]_p^{(i,j)}\right] = \rho \cdot \prod_{l=k}^{1} \frac{1}{\widetilde{\deg}(v_p^l)} \cdot w_q^{(l)}$. It follows that $\mathbb{E}\left[\frac{\partial h_x^{(k)}}{\partial h_y^{(0)}}\right] = \rho \cdot \prod_{l=k}^{1} W_l \cdot \left(\sum_{p=1}^{\Psi} \prod_{l=k}^{1} \frac{1}{\widetilde{\deg}(v_p^l)}\right)$. We know that the $k$-step random walk probability at $y$ can be computed by summing up the probability of all paths of length $k$ from $x$ to $y$, which is exactly $\sum_{p=1}^{\Psi} \prod_{l=k}^{1} \frac{1}{\widetilde{\deg}(v_p^l)}$. Moreover, the random walk probability starting at $x$ to other nodes sum up to 1. We know that the influence score $I(x, z)$ for any $z$ in expectation is thus the random walk probability of being at $z$ from $x$ at the $k$-th step, multiplied by a term that is the same for all $z$. Normalizing the influence scores ends the proof.

Comment: ReLU is not differentiable at 0. For simplicity, we assume the (sub)gradient to be 0 at 0. □

## B. Proof for Proposition 1

*Proof.* Let $\left[h_x^{(final)}\right]_i$ be the i-th entry of $h_x^{(final)}$, the feature after layer aggregation. For any node $y$, we have

$$I(x, y) = \sum_i \left\|\frac{\partial \left[h_x^{(final)}\right]_i}{\partial h_y^{(0)}}\right\|_1$$

$$= \sum_i \left\|\frac{\partial \left[h_x^{(j_i)}\right]_i}{\partial h_y^{(0)}}\right\|_1$$

where $j_i = \underset{l}{\arg\max}\left(\left[h_x^{(l)}\right]_i\right)$. By Theorem 1, we have

$$\mathbb{E}\left[I(x, y)\right] = \sum_l c_x^l \cdot z_l \cdot \mathbb{E}\left[I_x(y)^{(l)}\right]$$

where $I_x(y)$ is the $l$-step random walk probability at $y$, $z_l$ is a normalization factor and $c_x^l$ is the fraction of entries of $h_x^{(l)}$ being chosen by max-pooling. By Theorem 1, $\mathbb{E}\left[I_x(y)^{(l)}\right]$ is equivalent to the $l$-step random walk probability at $y$ starting at $x$.

□

## C. Visualization Results

We describe the details of the heat maps and present more visualization results. The colors of the nodes in the heat maps correspond to their probability masses of either the influence distribution or random walk distribution as shown in Figure 6. As we see, the shallower the color is, the smaller the probability mass. We use the same color for probabilities over 0.2 for better visual effects because there are few nodes with influence probability masses over 0.2. Nodes with probability mass less than 0.001 are not shown in the heat maps.

In Table 6, we present more visualization results to compare the 1) influence distributions under GCNs and the random walk distributions, 2) influence distributions under GCNs with residual connections and lazy random walk distributions. The nodes being influenced and the random walk starting node are labeled square. The influence distributions for the nodes in Figure 6 are computed according to Definition 3.1, under the same trained GCN (Res) models with 2, 4, 6 layers respectively. We use the hyper-parameters as described in Kipf & Welling (2017) for training the models. The graph (dataset) is taken from the Cora citation network as described in section 6. We compute the random walk distributions according to Definition 3.2 on the graph $\widetilde{G}$. The lazy ranndom walks all share the same lazy factor 0.4, i.e. there's an extra 0.4 probability of staying at the current node at each step. This probability is chosen for visual comparison with the GCN-ResNet. Note that the GCN and random walk colors may differ for nodes that have particularly large degrees because the models we are running follow Equation 2, which assigns less weight to nodes that have larger degrees, rather than Equation 3. The visualization in Figure 5 has the same setting as mentioned above. It is trained for the Cora dataset with a 6-layer JK Net with maxpooling layer aggregation.

| 2 layers / steps | | 4 layers / steps | | 6 layers / steps | |
|---|---|---|---|---|---|
| GCN / r.w. | Res / lazy r.w. | GCN / r.w. | Res / lazy r.w. | GCN / r.w. | Res / lazy r.w. |

Table 6. Influence distributions for (more) nodes under GCN, GCN-Res, and random walk distributions

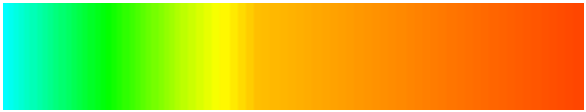

Figure 6. Color and probability correpondency for the heat maps

Next, we demonstrate subgraph structures where GCN models with 2 layers tend to make misclassification, whereas models with 3 or 4 layers are able to make the correct prediction and vice versa, with real dataset. These visualization results further complement and support the theory illustrated in Figure 1 and Theorem 1. As we see in Figure 7, a model with pre-fixed effective range priors, which looks at 2-hop neighbors, tends to make incorrect prediction if the local subgraph structure is tree-like (bounded treewidth). Thus, it would be desirable to look beyond the direct neighbors and draw information from nodes that are 3 or 4 hops away so as to learn a better representation of the local community. On the other hand, as we see in Figure 8, a model with pre-xied effective range priors, which looks at 3 or 4-hop neighbors, may happen to draw much information from less relevant neighbors and thus cannot learn the right representations, which are necessary for the correct prediction. In the subgraph structures where the random walk expansion explodes rapidly, models with 3 or 4 prefixed layers are essentially taking into account every node. Such global representations might not be ideal for the prediction for the node. In another scenario, despite possessing the locally bounded treewidth structure, because of the "bridge-like" structures, looking at distant nodes might imply drawing information from a completely different community, which would act like noises and influence the prediction results.

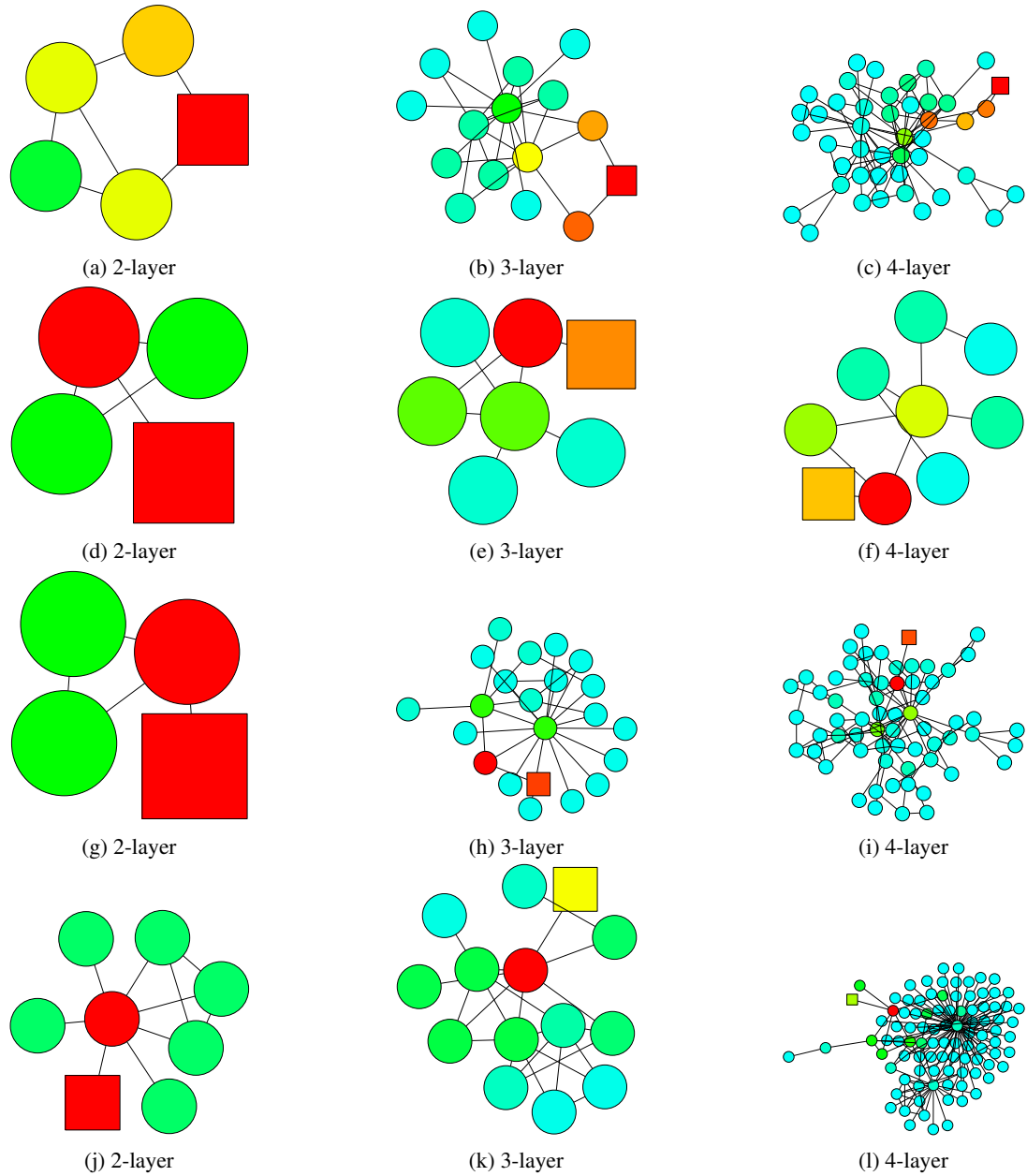

Figure 7. Subgraph structures where 2-layer GCNs make misclassification, whereas 3 and 4-layer GCNs make the correct prediction.

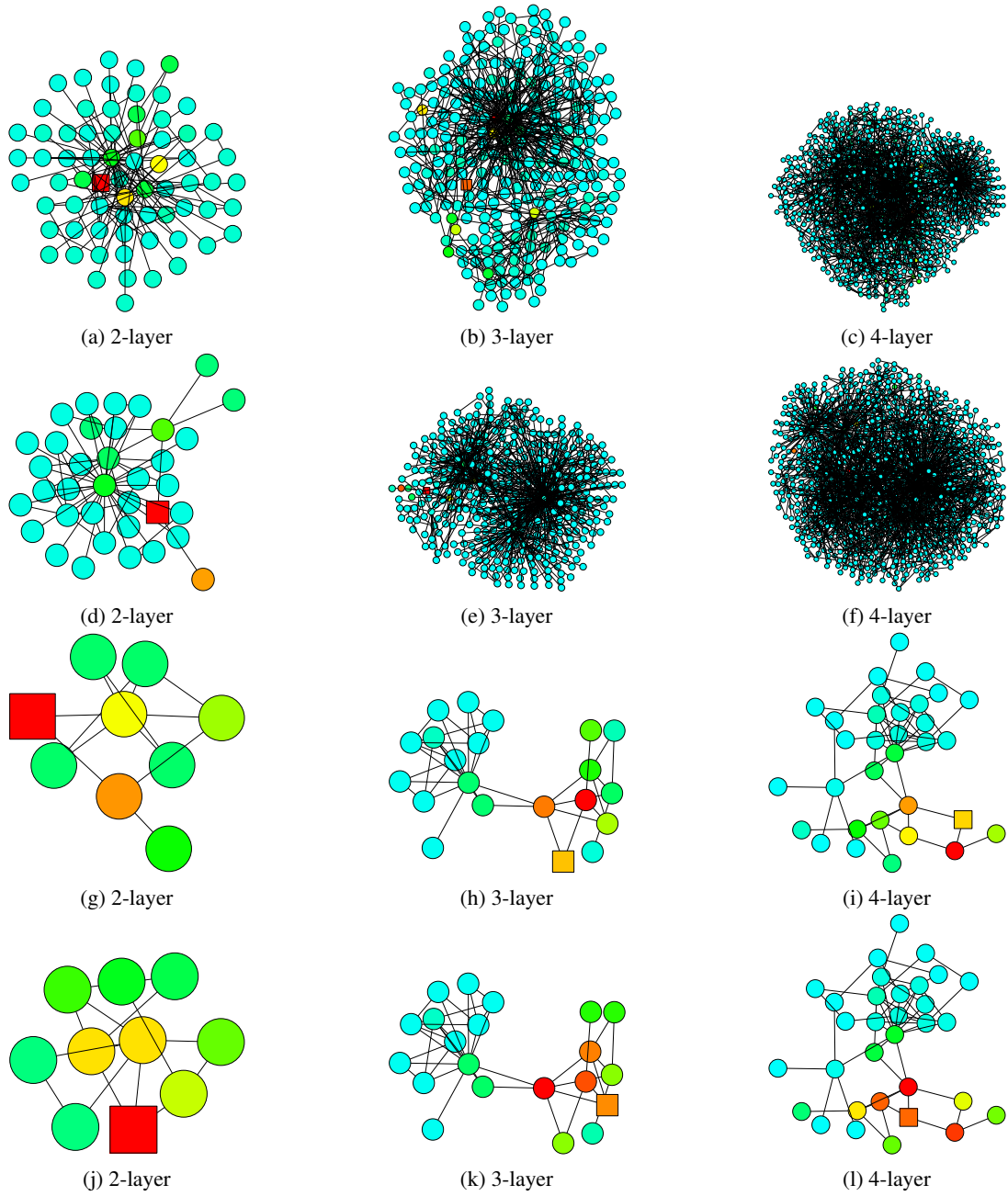

Figure 8. Subgraph structures where 3, 4-layer GCNs make misclassification, whereas 2-layer GCNs make the correct prediction.